\pdfoutput=1
\documentclass[11pt]{article}

\usepackage[final]{ACL2023}

\usepackage{times}
\usepackage{latexsym}

\usepackage[T1]{fontenc}
\usepackage[utf8]{inputenc}
\usepackage{amsmath,amssymb}
\usepackage{booktabs,multirow,xcolor}
\usepackage[colorinlistoftodos]{todonotes}
\usepackage{microtype}

\usepackage{inconsolata}

\title{Decomposing and Predicting Annotation Disagreement \\in Harmful Content Across Languages}

\author{
Irem Ecem Yelkanat$^{1\dagger}$ \qquad Faeze Ghorbanpour$^{1,2\dagger}$ \qquad Alexander Fraser$^{1,2}$\vspace{.2cm}\\
$^{1}$School of Computation, Information and Technology, TU Munich \\
$^{2}$Munich Center for Machine Learning (MCML)\\
\vspace{.1cm}
{\tt \small \{iremecem.yelkanat, faeze.ghorbanpour\}@tum.de} \\ 
{\small $^{\dagger}$Equal contribution.}
}

\begin{document}
\maketitle
\begin{abstract}
% Annotator disagreement is increasingly recognized as a meaningful property of subjective tasks, rather than merely noise to be removed. Predicting which items are likely to be contested could help allocate annotation effort and route difficult cases to human review. We propose predicting disagreement from interpretable semantic properties of the text. An LLM answers a set of binary questions covering harmful-content characteristics, disagreement-related cues such as borderline severity and missing context, and language-sensitive cues such as local slurs. A classifier predicts disagreement from these answers, and an explainer identifies which questions contribute most strongly to the predictions. Across four datasets in English, German, and Turkish and three open-weight LLMs, structured question features outperform direct prompting by 10.4 ROC-AUC points and a supervised text baseline by 3.6 points. A compact subset of eight questions performs within 0.7 ROC-AUC points of the full twenty-question set. In leave-one-dataset-out evaluation, models retain roughly 92–96\% of their in-domain ROC-AUC. Across models and datasets, questions about derogation, missing context, and decision boundaries serve as the strongest predictors, suggesting that what makes harmful content contested is partly observable in the text.
Annotator disagreement is increasingly recognized as a meaningful property of subjective tasks, rather than merely noise to be removed. Predicting which items are likely to be contested could help allocate annotation effort and route difficult cases to human review. We propose predicting disagreement from interpretable semantic properties of the text. An LLM answers a set of binary questions covering harmful-content characteristics, disagreement-related cues such as borderline severity and missing context, and language-sensitive cues such as local slurs. A classifier predicts disagreement from these answers, and we analyze which questions contribute most to the predictions. Across four datasets in English, German, and Turkish and three open-weight LLMs, our structured question features outperform direct prompting by 10.5 ROC-AUC points and the strongest text-based baselines by 5.0 points. A compact subset of eight questions performs within 1.5 ROC-AUC points of the full twenty-question set. In leave-one-dataset-out evaluation, models retain roughly 90–95\% of their in-domain ROC-AUC. Across models and datasets, questions about derogation, missing context, and decision boundaries serve as the strongest predictors, suggesting that what makes harmful content contested is partly observable in the text.

\end{abstract}

\section{Introduction}

Annotators often disagree when labeling subjective tasks. A post that one annotator marks as hateful, another may consider merely offensive, and a third may see as acceptable. For a long time, such disagreement was treated as noise, resolved through majority voting or adjudication, and then discarded~\citep{10.1162/tacl_a_00293}. A growing body of work now argues that disagreement is a meaningful property of subjective tasks, reflecting ambiguity, missing context, and differences in annotators' backgrounds and beliefs \citep{ plank-2022-problem, sap-etal-2022-annotators, davani2022dealing}. Harmful content is a clear case: what counts as hateful or offensive depends on who is reading, which cultural and linguistic norms apply, and how much context is available~\citep{pachinger-etal-2025-disaggregated}.

% If disagreement is informative, then predicting it is useful. A model that flags which items are likely to be contested could guide where to collect additional annotations, which items to route to human moderators rather than automatic decisions, and which parts of a dataset deserve closer auditing. In practice, however, disagreement labels are expensive, since they require several annotators per item. A disagreement predictor that works for new data or for datasets in other languages would reduce this cost.

If disagreement is informative, predicting it is useful. A disagreement predictor can indicate where to collect additional annotations, which items to route to human moderators~\citep{villate2024collaborative}, and which parts of a dataset need to be audited. Obtaining disagreement labels, however, requires several annotators per item, which is costly and rarely feasible for new languages. A natural shortcut is to use LLMs, which are now widely used as annotators \citep{tan-etal-2024-large, gligoric-etal-2025-unconfident}. 
%Yet LLMs capture the majority labels far better than human variation: 
However, they struggle to predict disagreement directly~\citep{ni-etal-2026-reasoning}
%, are overconfident on low-agreement items~\citep{lu-etal-2025-llm}, perform poorly on tasks without consensus (Radharapu et al., 2025)
, and tend to collapse the diversity of human judgments~\citep{chochlakis-etal-2025-humans}. Specifically for harmful content, LLMs align more with the Western community, even when prompted in the native language \citep{bui-etal-2025-multi3hate, masud-etal-2024-hate}.

Existing approaches leave two gaps. First, they offer limited insight into why an item is contested. One line of work models disagreement from the annotator side, using their identities, demographics, or beliefs \citep{davani2022dealing, fleisig-etal-2023-majority, sap-etal-2022-annotators}. This requires annotator-level information that is often unavailable, and demographics alone explain only part of the variation~\citep{orlikowski-etal-2023-ecological}. Another line trains classifiers to predict disagreement directly from text~\citep{fornaciari-etal-2021-beyond, villate2024collaborative}, or aligns LLM uncertainty with human disagreement through fine-tuning~\citep{lu2026aligningllmuncertaintyhuman}. These models can be accurate, but their predictions are not tied to explicit properties of the text. Analyses of text-side factors exist~\citep{sandri-etal-2023-dont, falk-lapesa-2025-mining}, but they rely on manual annotation or post hoc correlation rather than on a predictive model. Second, the obvious alternative of asking an LLM whether annotators would disagree works poorly. LLMs struggle to capture annotator disagreement~\citep{ni-etal-2026-reasoning}, and they tend to be overconfident on the low-agreement items that matter most~\citep{lu-etal-2025-llm}. 
%Finally, nearly all of this work is monolingual, so it remains unclear whether disagreement patterns learned on one dataset or language carry over to another.

% We propose predicting disagreement from interpretable properties of the text. Rather than asking an LLM for a holistic judgment, we ask a set of targeted binary questions about each item. The questions fall into three groups: questions about derogatory and hateful content (whether the text uses slurs or dehumanizes a group), questions about boundary cues associated with disagreement (whether the severity is borderline or context is missing), and questions about language- and culture-specific cues (whether the text contains a local slur or coded insult). A random forest classifier predicts disagreement from the answers, and we use TreeSHAP to attribute each prediction to individual questions. Unlike token attributions or a direct LLM judgment, our features are interpretable by construction: each is the answer to a named question, so importance can be read at the level of concepts rather than words.
We propose predicting disagreement from interpretable properties of the text. Rather than asking an LLM for a holistic judgment, we ask a set of targeted binary questions about each item. The questions fall into three groups: questions about derogatory and hateful content (whether the text uses slurs or dehumanizes a group), questions about boundary cues associated with disagreement (whether the severity is borderline or context is missing), and questions about language- and culture-specific cues (whether the text contains a local slur or coded insult). A random forest classifier predicts disagreement from the answers, and we use TreeSHAP to attribute each prediction to individual questions. Unlike token attributions or a direct LLM judgment, our features are interpretable by construction: each is the answer to a named question, so importance can be read at the level of concepts rather than words. We evaluate this across three open-weight LLMs, Gemma 3 27B, Mistral Small 3.2 24B, and Qwen 3.6 27B, and four datasets spanning three languages: the English MHS and HateXplain, the Turkish Hate Speech Dataset, and the German DeTox. 

% We evaluate on four harmful content datasets in English, German, and Turkish, using three open-weight LLMs. %, and ask three research questions:
% RQ1: Can interpretable question-based features predict annotator disagreement, and how do they compare with direct LLM prompting and supervised text baselines?
% RQ2: Which questions drive the predictions, and can a small subset of questions retain the predictive signal?
% RQ3: Do disagreement predictors trained on some datasets and languages transfer to unseen ones?
Our contributions are:
\textbf{(1):} We formulate annotator disagreement prediction as answering interpretable questions about the text, and introduce a set of twenty questions covering harmful content, boundary cues, and language-specific cues.
\textbf{(2):} We show across four datasets, three languages, and three LLMs that these features predict disagreement better than direct prompting and text-based baselines, and that eight questions suffice.
\textbf{(3):} We analyze which questions drive disagreement predictions and release the questions, prompts, and LLM answers for all datasets and show that question-based disagreement predictors transfer across datasets, retaining most of their in-domain performance.
% \textbf{(4):} 

Our question-based approach outperforms direct prompting by 10.5 ROC-AUC points and the strongest supervised embedding baseline by 5.0 points. Boundary and context questions add a consistent signal on top of content questions, and a subset of eight questions performs within 1.5 points of the full set of twenty, reducing LLM calls by 60\%. In leave-one-dataset-out evaluation, without any explicit language or dataset feature, models retain 90–95\% of their in-domain ROC-AUC. Questions on derogation and context are the most consistently important across languages and models.

\section{Related work}
\paragraph{Annotator Disagreement in Subjective Tasks}
Disagreement in subjective annotation was long treated as noise, but a growing body of work shows it is systematic. \citet{akhtar2020modeling} found polarized annotator groups in hate speech, \citet{leonardelli-etal-2021-agreeing} showed that offensive language datasets contain many low-agreement items, and \citet{plank-2022-problem} framed label variation as a property of the data to be modeled. Later work examined why annotators disagree. \citet{sap-etal-2022-annotators} linked toxicity ratings to annotator beliefs and identities, \citet{sandri-etal-2023-dont} proposed a taxonomy of disagreement reasons such as ambiguity and missing context, and \citet{pachinger-etal-2025-disaggregated} showed that high-disagreement toxic items reflect subjective perception rather than inconsistency. Disagreement also has a cultural dimension, as shown for English hate speech \citep{lee-etal-2024-exploring-cross}, multilingual irony \citep{casola-etal-2024-multipico}, and multilingual hateful memes \citep{bui-etal-2025-multi3hate}. These studies identify the sources of disagreement through manual analysis or correlation \citep{falk-lapesa-2025-mining}, but none use these sources to predict disagreement. Our boundary and context questions operationalize these reasons as automatically extracted features.

\paragraph{Modeling and Predicting Disagreement}

Early models of disagreement took the annotator's perspective, predicting individual labels with per-annotator heads \citep{davani2022dealing} or from demographics \citep{fleisig-etal-2023-majority}. These need annotator metadata, which explains only part of the variation \citep{orlikowski-etal-2023-ecological}. Text-side approaches instead learn label distributions, from soft-label multi-task learning \citep{fleisig-etal-2023-majority} to disagreement as an auxiliary moderation task \citep{villate2024collaborative} and LLMs fine-tuned to express human-like uncertainty \citep{lu2026aligningllmuncertaintyhuman}. Their predictions, however, are not tied to explicit properties of the text. More recently, LLMs have been tested as direct judges of disagreement, and they fall short: they reproduce majority views \citep{chochlakis-etal-2025-humans}, are overconfident on low-agreement offensive items \citep{lu-etal-2025-llm}, and struggle to predict disagreement when asked directly \citep{ni-etal-2026-reasoning}. We avoid both limitations. LLMs answer targeted questions they can judge reliably, and an interpretable model learns the mapping to disagreement, without annotator metadata.

\paragraph{Interpretable Harmful Content Detection}
Interpretability for harmful content began with token-level rationales, such as HateXplain \citep{mathew2022hatexplainbenchmarkdatasetexplainable}, and post-hoc attribution methods, which can be unfaithful \citep{10.1145/3375627.3375830} and highlight words rather than concepts. Concept bottleneck models \citep{koh2020concept} address this by routing predictions through human-readable concepts, an approach that has recently been scaled to LLM-based text classification \citep{sun2025concept}. For harmful content, LLM-scored adjectives have served as concepts for hate and counter-speech detection \citep{labadie2026distilling}, checklists of binary questions answered by an LLM have similarly been used to make hate speech classification interpretable \citep{girón2026xlisthatechecklistbasedframeworkinterpretable}, and LLM-rated hate attributes recover aggregate hate scores more effectively than direct prompting \citep{jradi2026attributebaseddiagnosisllmalignment}. All of these target the harmfulness label. We apply concept-based prediction to a different target: annotator agreement, which requires concepts that capture what makes content contested rather than what makes it harmful.

\paragraph{Cross-Lingual and Cross-Dataset Transfer}
Harmful content classifiers transfer poorly. Zero-shot cross-lingual detection misses language-specific expressions \citep{nozza-2021-exposing}, target-language data remains necessary \citep{rottger2022data, ghorbanpour-etal-2025-fine}, classifiers are culturally insensitive \citep{lee-etal-2023-hate}, and LLM moderation shifts with geographic cues \citep{masud-etal-2024-hate}. Recent work narrows the gap through multilingual retrieval \citep{ghorbanpour-etal-2025-data}. Disagreement modeling, by contrast, has been evaluated almost exclusively within a single dataset or language \citep{dehghan2025dealing, leonardelli-etal-2025-lewidi, zaghouani2026annotation}, likely because transferring disagreement predictors to unseen datasets and languages is more challenging than label transfer, as what is considered contested can depend on local norms. Because our questions are defined independently of any dataset or language, the same representation applies across languages, allowing direct leave-one-dataset-out evaluation.

% Requires in preamble: \usepackage{amsmath,amssymb}

\section{Methodology}
\label{sec:method}

We predict annotator disagreement from structured properties of the text. For each item, an LLM answers a fixed set of binary questions covering harmful content, boundary cues, and language-specific cues, and we extract the model's confidence in each answer from its token probabilities. The answers, confidences, and pairwise combinations of answers form a feature vector that is shared across datasets. A random forest trained on this vector predicts whether annotators disagree, both within a dataset and on unseen datasets.

We use a deliberately minimal, dataset-agnostic target: whether at least two annotators assign different
labels. So defined, it is invariant to the label inventory and depends only on label identity, not meaning, so binary and multi-class schemes, and
intermediate labels, are treated uniformly. Observed disagreement does rise with the number of annotators per item, but across our datasets, disagreement rates do not track annotator counts, indicating that count is not the dominant factor. We also train and evaluate within each dataset rather than comparing raw disagreement magnitudes across them, so differences in annotation protocol do not enter our claims.
% A count-robust, graded target and a matched-count analysis (Limitations) would isolate these effects further.

\subsection{Problem Formalization}
\label{sec:formalization}

Let $d \in D$ be an item from a dataset $D$, annotated by $n_d$ annotators. Each annotator $i$ assigns a label $a_{d,i} \in L$, giving the annotation set $A_d = \{a_{d,1}, \dots, a_{d,n_d}\}$. The label set $L$ depends on the dataset.
%We keep intermediate labels such as \textit{offensive} distinct, so an item labeled \textit{hate} by one annotator and \textit{offensive} by another counts as contested. 
The disagreement label of $d$ is
\begin{equation}
    y_d = \mathbb{1}\big[\, |\{a \mid a \in A_d\}| > 1 \,\big],
    \label{eq:target}
\end{equation}
so $y_d = 1$ if at least two annotators assign different labels and $y_d = 0$ otherwise. Given only the text of $d$, a model predicts a disagreement score $\hat{y}_d \in [0, 1]$, without access to annotator identities, demographics, or individual labels.

We consider two settings over datasets $\mathcal{D} = \{D_1, \dots, D_K\}$ that differ in language, label set, and annotation guidelines. In the \textbf{in-domain} setting, a model is trained and evaluated on the same $D_k$. In the \textbf{transfer} setting, it is trained on $\mathcal{D} \setminus \{D_k\}$ and evaluated on $D_k$, without information identifying the language or source of the data.

\subsection{Disagreement Prediction}
\label{sec:prediction}

% \paragraph{Question answering.}
Let $Q = Q_C \cup Q_B \cup Q_L$ be a set of binary questions about observable properties of the text. \textit{Content questions} ($Q_C$) capture whether and how the text is harmful, such as references to protected groups, slurs, or dehumanization. \textit{Boundary questions} ($Q_B$) capture properties that make the label hard to decide, such as an ambiguous target, borderline severity, missing context, or sarcasm. \textit{Language and context questions} ($Q_L$) capture cues that require local knowledge, such as local slurs, coded insults, and community-specific pragmatic cues. For each item $d$ and question $q_j \in Q$, we prompt an LLM $M$ with the question and the text, and obtain a binary answer
\begin{equation}
    s_{d,j} = M(q_j, d) \in \{0, 1\}.
    \label{eq:answer}
\end{equation}

% \paragraph{Confidence extraction.}
We derive the confidence of each answer from the token-level log probabilities of the two answer tokens $\mathcal{V} = \{\textit{yes}, \textit{no}\}$, renormalized with a softmax:
\begin{equation}
    c_{d,j} = \frac{\exp(\ell_{\hat{k}})}{\sum_{k \in \mathcal{V}} \exp(\ell_k)},
    \label{eq:confidence}
\end{equation}
where $\ell_k$ is the log probability of token $k$ and $\hat{k}$ is the generated answer token. A low $c_{d,j}$ indicates that the LLM is uncertain whether the property is present.

% \paragraph{Feature construction.}
Disagreement often depends on combinations of properties. Derogatory content that is also borderline in severity may be contested even when neither property alone predicts disagreement. We therefore represent each item by its answers and the corresponding confidences:
\begin{equation}
    \mathbf{x}_d = \big[\, s_{d,j} \,\big]_{j} \oplus \big[\, c_{d,j} \,\big]_{j},
    \label{eq:features}
\end{equation}
% \begin{equation}
%     \mathbf{x}_d = \big[\, s_{d,j} \,\big]_{j} \oplus \big[\, c_{d,j} \,\big]_{j} \oplus \big[\, s_{d,j}\, s_{d,k} \,\big]_{j<k},
%     \label{eq:features}
% \end{equation}
where $\oplus$ denotes concatenation. We also experimented with pairwise products of answers, but they did not improve performance and are omitted.

% \paragraph{Prediction.}
\noindent
A random forest predicts the disagreement score,
\begin{equation}
    \hat{y}_d = \mathrm{RF}(\mathbf{x}_d),
    \label{eq:prediction}
\end{equation}
and is trained on items with observed labels $y_d$. Because $\mathbf{x}_d$ is defined over the same questions for every dataset, it does not depend on the language or label set of the data. The same model therefore applies to both settings: in-domain it is trained on $D_k$, and in transfer it is trained on $\mathcal{D} \setminus \{D_k\}$ and applied to $D_k$ directly.

\section{Experimental Details}

\paragraph{Datasets.} We use four harmful-content datasets that provide labels from multiple annotators per item. MHS (Measuring Hate Speech)
\citep{kennedy2026measuringhatespeechspectrum, sachdeva-etal-2022-measuring} contains English comments from Reddit, Twitter, YouTube, and Gab, annotated by a large pool of crowdworkers with detailed demographic records; we use $29{,}453$ comments, of which $40.4\%$ have disagreement. HateXplain \citep{mathew2022hatexplainbenchmarkdatasetexplainable} contains English posts from Twitter and Gab, each labeled by three annotators as hate speech, offensive, or normal; we use $19{,}192$ comments, with a disagreement rate of $48.7\%$. The Turkish Hate Speech Dataset \citep{toraman-etal-2022-large} contains Turkish tweets across several target domains, such as religion, gender, race, politics, and sports, labeled as hate, offensive, or normal; we use the $60{,}308$ multiply-annotated comments, of which $46.6\%$ have disagreement. DeTox \citep{demus-etal-2022-comprehensive} contains German comments from news and social media platforms, annotated by multiple annotators with a binary label; we use $9{,}689$ comments, with a disagreement rate of $21.5\%$. Further details are in Appendix~\ref{app:datasets}.

% For the three datasets with an offensive category, disagreement includes cases where annotators choose between hate and offensive. The datasets therefore differ in language, source platform, label set, annotation guidelines, and number of annotators per item, which makes transfer between them a demanding test. 

\paragraph{Models.} We extract question answers with three open-weight instruction-tuned LLMs of similar size from different families: Gemma 3 27B \citep{gemmateam2025gemma3technicalreport}, Mistral Small 3.2 24B \citep{mistral_small_32_2025}, and Qwen 3.6 27B \citep{qwen3.6-27b}. Using several families lets us test whether results depend on a specific model. All LLMs are used for inference only, with greedy decoding, and generation is limited to a single answer token so that token probabilities can be read directly. As the disagreement classifier, we use a random forest with balanced class weights to account for the different disagreement rates across datasets. A separate classifier is trained for each LLM, so features from different LLMs are never mixed. Inference settings and classifier hyperparameters are in Appendix~\ref{app:models}.

\begin{table}[t]
% \centering
\small
\begin{tabular}{ll}
\toprule
\textbf{ID} & \textbf{Short description} \\
\midrule

\multicolumn{2}{p{7cm}}{\textit{Content questions (C)}} \\
q1  & Protected group referenced \\
q2  & Negative or derogatory description \\
q3  & Slurs or stereotypes \\
q4  & Dehumanization or vilification \\
q5  & Blaming or scapegoating \\
q6  & Exclusion or discrimination \\
q7  & Threats or wishes of harm \\
q8  & Direct incitement \\
q9  & Context or intent assessment \\
q10 & Likely effect on target \\
\midrule

\multicolumn{2}{p{7cm}}{\textit{Boundary questions (B)}} \\
d1 & Target ambiguity \\
d2 & Abuse-vs-harm boundary \\
d3 & Missing context \\
d4 & Slang, sarcasm, or coded language \\
d5 & Borderline severity \\
\midrule

\multicolumn{2}{p{7cm}}{\textit{Language and context questions (L)}} \\
m1 & Local slur or coded insult \\
m2 & Local protected-group wording \\
m3 & Local conflict frame \\
m4 & Indirect local exclusion \\
m5 & Local pragmatic cue \\
\bottomrule
\end{tabular}

\caption{Question IDs and short descriptions grouped into content, boundary, and language/context questions.}
\label{tab:question_descriptions}
\end{table}

\paragraph{Prompts.} Each question is asked in a separate prompt, so the answer to one question cannot influence another. Each prompt is few-shot, and the model answers with \emph{yes} or \emph{no}. The content questions are adopted from the checklist of \citet{girón2026xlisthatechecklistbasedframeworkinterpretable}, which decomposes hate speech into established definitional components. Every prompt has the same structure: a system instruction describing the role of a content analyst, a short definition of the property being asked about, the text to analyze, and an instruction to answer only \textit{yes} or \textit{no}. The definitions are shared across datasets and do not refer to any dataset's annotation guidelines or label set, keeping the representation independent of how each dataset was annotated. Prompts are written in English for all datasets, while the text is kept in its original language, so the LLM handles German and Turkish input with English instructions. %Content, boundary, and language and context questions use the same template and differ only in the definition. 
The full list of questions and prompt templates is in Appendix~\ref{app:prompts}.

\paragraph{Evaluation.} For in-domain experiments, we use 3-fold cross-validation grouped by normalized text, so that no text appears in more than one fold. Within each training fold, we hold out a validation split to tune the decision threshold. For transfer, we use leave-one-dataset-out evaluation: the classifier is trained on three datasets and tested on the fourth, with no explicit dataset or language feature given to the classifier. Our primary metrics are threshold-independent: ROC-AUC and PR-AUC, which are sensitive to class imbalance. We additionally report macro-F1 and MCC, selecting their thresholds on validation data. For transfer, we report relative ROC-AUC, the ratio of transfer to in-domain ROC-AUC on the same dataset. %Further details are in Appendix D.

\paragraph{Baselines.} We compare the question-based predictor against several baselines, all evaluated on the same instances and folds. Direct prompting is a label-free reference that asks the same LLM (few-shot) whether annotators would disagree and scores the item by its probability of "disagree." As text baselines, we train the same random forest used by our method on multilingual sentence embeddings (LaBSE~\citep{feng-etal-2022-language}, multilingual E5 base~\citep{wang2024multilinguale5textembeddings}, and Qwen 3 0.6B embeddings~\citep{qwen3embedding}), testing whether a text representation alone predicts disagreement. We fix the classifier to a random forest rather than fine-tuning an LLM, so any difference reflects the input representation rather than classifier capacity, and it supports per-question TreeSHAP attribution. We replaced the random forest model with XGBoost or logistic regression, and the results changed negligibly. We additionally report two privileged, label-based controls: a random forest trained on the gold majority label, which tests whether disagreement is explained by the label itself, and one trained on the gold label together with the LLM’s harmful-content output. These controls are privileged because the gold majority label is not available for a new, unlabeled item, and they appear in Appendix~\ref{app:label-controls}.

\begin{table*}[t]\centering\scriptsize\setlength{\tabcolsep}{2.6pt}
\begin{tabular}{@{}cl cccc cccc cccc cccc@{}}\toprule
& & \multicolumn{4}{c}{\textbf{MHS}} & \multicolumn{4}{c}{\textbf{HateXplain}} & \multicolumn{4}{c}{\textbf{Turkish}} & \multicolumn{4}{c}{\textbf{DeTox}} \\
\cmidrule(lr){3-6}\cmidrule(lr){7-10}\cmidrule(lr){11-14}\cmidrule(lr){15-18}
& & ROC & PR & F1 & MCC & ROC & PR & F1 & MCC & ROC & PR & F1 & MCC & ROC & PR & F1 & MCC \\\midrule
\multirow{11}{*}{\rotatebox[origin=c]{90}{\textbf{Gemma 3 27B}}}
& Direct prompting   & 63.8 & 50.5 & 50.8 & 23.2 & 58.1 & 54.4 & 43.8 & 14.0 & 68.0 & 58.9 & 58.9 & 34.0 & 64.3 & 27.3 & 35.7 & 19.2 \\
& LaBSE              & 68.3 & 53.8 & 62.2 & 24.9 & 60.2 & 57.8 & 57.2 & 14.6 & 69.6 & 61.2 & 64.0 & 29.9 & 74.6 & 41.3 & 63.9 & 28.6 \\
& mE5                & 70.1 & 56.2 & 63.7 & 28.4 & 60.1 & 57.8 & 57.1 & 14.4 & 69.7 & 61.5 & 63.9 & 29.1 & 75.6 & 42.2 & 64.3 & 30.3 \\
& Qwen-emb           & 70.2 & 56.3 & 63.7 & 28.5 & 61.2 & 59.1 & 57.7 & 15.9 & 67.9 & 60.1 & 62.6 & 26.1 & 74.5 & 42.7 & 64.2 & 28.6 \\
& C                  & 70.6 & 44.7 & 60.9 & 30.0 & 60.0 & 55.2 & 56.1 & 13.9 & 68.7 & 61.0 & 65.2 & 32.7 & 76.0 & 40.7 & 63.0 & 36.3 \\
& C + B              & 71.3 & 45.6 & 61.0 & 30.1 & 62.2 & 57.5 & 58.0 & 17.4 & 71.3 & 62.3 & 65.6 & 35.9 & 77.7 & 41.0 & 63.6 & 36.7 \\
& C + L              & 71.4 & 46.9 & 61.0 & 30.4 & 61.1 & 56.7 & 56.0 & 14.5 & 70.0 & 61.7 & 65.5 & 34.3 & 77.9 & 42.0 & 64.1 & 36.2 \\
& C + B + L          & 71.8 & 47.0 & 61.2 & 30.3 & 63.3 & 58.9 & 58.3 & 18.4 & 71.8 & 62.8 & 65.8 & \textbf{36.7} & 78.6 & 42.3 & 63.3 & 36.5 \\
& C + B + L + conf.  & \textbf{74.7} & \textbf{61.3} & \textbf{67.9} & \textbf{37.0} & \textbf{64.6} & \textbf{60.9} & \textbf{60.4} & \textbf{21.0} & \textbf{73.4} & \textbf{64.7} & \textbf{67.3} & \underline{36.2} & \textbf{81.6} & \textbf{48.7} & \textbf{69.7} & \textbf{41.2} \\
& 8 questions        & 71.3 & 46.1 & 60.5 & 30.6 & 62.3 & 57.8 & 58.4 & 17.3 & 70.4 & 61.0 & 65.4 & 34.5 & 77.9 & 40.6 & 63.5 & 36.6 \\
& 8 questions + conf. & \underline{73.8} & \underline{60.1} & \underline{67.0} & \underline{35.1} & \underline{63.8} & \underline{60.2} & \underline{59.4} & \underline{18.9} & \underline{71.8} & \underline{62.8} & \underline{66.1} & 34.9 & \underline{80.2} & \underline{46.6} & \underline{68.1} & \underline{39.0} \\
\cmidrule(l){2-18}
\multirow{11}{*}{\rotatebox[origin=c]{90}{\textbf{Mistral Small 3.2 24B}}}
& Direct prompting   & 59.6 & 47.5 & 55.6 & 11.9 & 59.0 & 56.2 & 55.7 & 11.8 & 67.3 & 58.7 & 62.8 & 26.9 & 70.6 & 33.6 & 59.5 & 21.4 \\
& LaBSE              & 68.3 & 53.8 & 62.2 & 24.9 & 60.2 & 57.8 & 57.2 & 14.6 & 69.6 & 61.2 & 64.0 & 29.9 & 74.6 & 41.3 & 63.9 & 28.6 \\
& mE5                & 70.1 & 56.2 & 63.7 & 28.4 & 60.1 & 57.8 & 57.1 & 14.4 & 69.7 & 61.5 & 63.9 & 29.1 & 75.6 & 42.2 & 64.3 & 30.3 \\
& Qwen-emb           & 70.2 & 56.3 & 63.7 & 28.5 & 61.2 & 59.1 & 57.7 & 15.9 & 67.9 & 60.1 & 62.6 & 26.1 & 74.5 & 42.7 & 64.2 & 28.6 \\
& C                  & 69.5 & 47.5 & 62.5 & 28.2 & 60.9 & 58.4 & 58.1 & 16.1 & 62.4 & 56.8 & 61.1 & 22.2 & 60.5 & 29.6 & 60.6 & 21.2 \\
& C + B              & 71.6 & 49.1 & 62.1 & 28.1 & 63.1 & 59.2 & 59.4 & 18.9 & 69.3 & 60.8 & 64.1 & \underline{33.4} & 76.1 & 37.6 & 60.0 & 34.9 \\
& C + L              & 71.0 & 48.7 & 62.6 & 29.9 & 61.7 & 58.7 & 58.6 & 17.7 & 65.7 & 59.1 & 62.4 & 25.0 & 71.5 & 36.1 & 60.7 & 29.9 \\
& C + B + L          & 72.2 & 49.9 & 62.9 & 29.9 & 63.4 & 59.4 & 59.6 & 19.5 & 70.0 & 61.2 & 64.5 & 33.3 & 77.1 & 39.6 & 61.4 & 34.1 \\
& C + B + L + conf.  & \textbf{76.1} & \underline{63.1} & \textbf{69.2} & \textbf{39.3} & \textbf{65.5} & \textbf{61.7} & \textbf{60.9} & \textbf{22.4} & \textbf{72.2} & \textbf{63.5} & \textbf{66.3} & \textbf{35.2} & \textbf{82.4} & \textbf{50.4} & \textbf{69.6} & \textbf{40.5} \\
& 8 questions        & 71.4 & 48.3 & 63.0 & 29.7 & 61.9 & 58.7 & 58.1 & 16.3 & 67.7 & 59.8 & 64.3 & 32.1 & 74.5 & 37.4 & 59.6 & 34.7 \\
& 8 questions + conf. & \underline{75.8} & \textbf{63.2} & \underline{68.6} & \underline{37.8} & \underline{64.9} & \underline{61.0} & \underline{60.6} & \underline{21.2} & \underline{71.2} & \underline{62.3} & \underline{65.4} & 33.0 & \underline{81.2} & \underline{47.6} & \underline{69.1} & \underline{40.2} \\
\cmidrule(l){2-18}
\multirow{11}{*}{\rotatebox[origin=c]{90}{\textbf{Qwen 3.6 27B}}}
& Direct prompting   & 59.9 & 45.9 & 55.7 & 13.7 & 59.1 & 56.2 & 55.5 & 11.3 & 67.1 & 58.4 & 62.2 & 26.2 & 67.5 & 30.9 & 57.8 & 16.9 \\
& LaBSE              & 68.3 & 53.8 & 62.2 & 24.9 & 60.2 & 57.8 & 57.2 & 14.6 & 69.6 & 61.2 & 64.0 & 29.9 & 74.6 & 41.3 & 63.9 & 28.6 \\
& mE5                & 70.1 & 56.2 & 63.7 & 28.4 & 60.1 & 57.8 & 57.1 & 14.4 & 69.7 & 61.5 & 63.9 & 29.1 & 75.6 & 42.2 & 64.3 & 30.3 \\
& Qwen-emb           & 70.2 & 56.3 & 63.7 & 28.5 & 61.2 & 59.1 & 57.7 & 15.9 & 67.9 & 60.1 & 62.6 & 26.1 & 74.5 & 42.7 & 64.2 & 28.6 \\
& C                  & 71.0 & 47.6 & 63.2 & 31.1 & 61.6 & 58.0 & 58.3 & 17.7 & 65.3 & 59.3 & 63.0 & 26.5 & 73.1 & 40.3 & 66.0 & 35.8 \\
& C + B              & 72.2 & 49.4 & 62.6 & 30.8 & 62.9 & 59.0 & 58.6 & 18.3 & 69.4 & 62.0 & 64.4 & 33.2 & 78.8 & 45.0 & 63.7 & 35.6 \\
& C + L              & 71.8 & 49.4 & 62.8 & 30.7 & 62.5 & 59.2 & 58.4 & 17.6 & 67.3 & 61.0 & 63.6 & 27.5 & 76.7 & 42.1 & 64.4 & 35.0 \\
& C + B + L          & 72.9 & 50.8 & 63.6 & 31.5 & 63.3 & 59.6 & 59.0 & 18.4 & 70.1 & 62.7 & 64.7 & 33.2 & 79.3 & 45.2 & 64.2 & 36.9 \\
& C + B + L + conf.  & \textbf{76.3} & \underline{63.3} & \textbf{69.1} & \textbf{38.6} & \textbf{66.0} & \textbf{62.5} & \textbf{61.5} & \textbf{23.1} & \textbf{73.2} & \textbf{65.0} & \textbf{66.7} & \textbf{35.9} & \textbf{84.1} & \textbf{51.9} & \textbf{71.5} & \textbf{44.7} \\
& 8 questions        & 72.0 & 49.0 & 62.9 & 31.1 & 61.8 & 58.6 & 58.2 & 16.4 & 67.4 & 60.3 & 64.0 & 30.1 & 77.0 & 42.4 & 62.8 & 34.0 \\
& 8 questions + conf. & \underline{76.0} & \textbf{63.4} & \underline{69.0} & \underline{38.5} & \underline{64.7} & \underline{61.6} & \underline{59.6} & \underline{19.6} & \underline{71.7} & \underline{63.2} & \underline{65.8} & \underline{33.2} & \underline{82.5} & \underline{49.5} & \underline{69.9} & \underline{41.2} \\
\bottomrule\end{tabular}
\caption{Disagreement prediction (\%), three outer folds. LaBSE, mE5, and Qwen-emb are multilingual-embedding baselines. C, B, L are content, boundary, and language/context question groups; conf.\ adds answer confidences. Best per column within each LLM block in \textbf{bold}, second best \underline{underlined}. }
\label{tab:main}\end{table*}

\section{Results}
\label{sec:results}

We first evaluate how well question-based features predict disagreement and how much each question set contributes (\S\ref{sec:res-prediction}). We then analyze which questions drive the predictions (\S\ref{sec:res-explanation}) and if predictors transfer to unseen datasets and languages (\S\ref{sec:res-transfer}). 
% We report all four metrics: ROC-AUC, PR-AUC, macro-F1, and MCC averaged over 5-fold for each LLM.

\subsection{Predicting Disagreement}
\label{sec:res-prediction}

\paragraph{Comparison with baselines.}
Table~\ref{tab:main} presents the performance of the proposed method and compares it with the results from baseline studies. For all three LLMs, question-based features with confidence outperform the label-free baselines on every dataset. Asking the LLM directly whether annotators would disagree is the weakest approach: the same LLMs improve by 10.5 ROC-AUC points on average when asked targeted questions instead. The gains are largest on DeTox ($+15.2$) and MHS ($+14.6$), the two datasets with the lowest disagreement rates, and smallest on Turkish ($ +5.5$) and HateXplain ($ +6.6$). This is consistent with prior findings that LLMs struggle to estimate disagreement when asked directly \citep{ni-etal-2026-reasoning,lu2026aligningllmuncertaintyhuman}. Question features also outperform the embedding-based text baselines for all 12 dataset--LLM combinations, by $3.6$ ROC-AUC points on average. The full model performs similarly across LLMs, with differences of at most $2.5$ ROC-AUC points on any dataset, and Qwen is best on three of four datasets. We additionally compare against two privileged controls that take the human gold label as input; our question-based features outperform both on most datasets, the exception being HateXplain (Appendix~\ref{app:label-controls}). Across three outer folds, the standard deviations are small, below $0.5$ ROC-AUC points on the larger datasets and higher on HateXplain and DeTox, and in all cases smaller than the gaps between our model and the baselines, so the method ranking is stable (Appendix~\ref{app:std}).

\paragraph{Contribution of question groups.}
Boundary and language/context questions add signal beyond content questions for every LLM and dataset. Combining both groups with content questions improves ROC-AUC by $2.5$ points (Gemma), $7.3$ points (Mistral), and $3.7$ points (Qwen) on average. Boundary questions contribute more than language/context questions, with average gains of $1.8$, $6.7$, and $3.1$ points compared to $1.3$, $4.2$, and $1.8$, and a larger gain in 10 of 12 combinations. For Mistral and Qwen, the gains are largest on the non-English datasets: DeTox ($+16.6$ and $+6.2$) and Turkish ($+7.6$ and $ +4.8$), compared to at most 2.7 on the English datasets. For Gemma, gains are more uniform across datasets ($1.2$--$3.3$). Adding answer confidence further improves ROC-AUC by $1.4$, $2.4$, and $2.7$ points, again in all 12 combinations, indicating that how clearly a property is present is informative beyond whether it is present. Adding pairwise answer interactions instead of confidence yields a smaller gain, and adding them on top of confidence does not help further ($73.5$ vs.\ $73.6$ average ROC-AUC), so we use confidence only.

\begin{figure}[t]
\centering
\includegraphics[width=\columnwidth]{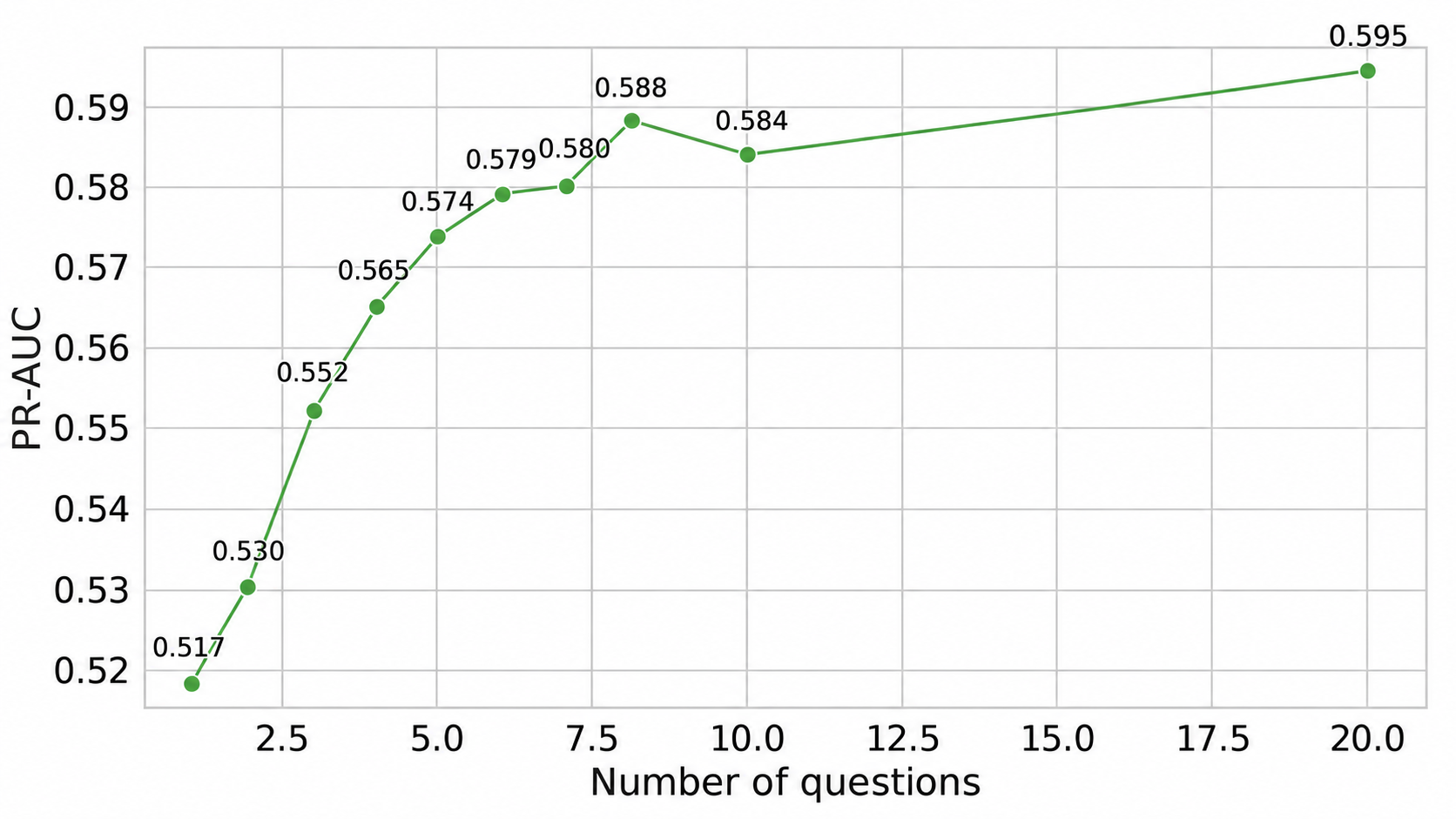}
\caption{Held-out test PR-AUC as a function of the number of selected questions.}
\label{fig:subset-curve}
\end{figure}

\paragraph{Compact question set.}
Since each question costs one LLM call per item, we select a compact subset. For each LLM, we run a beam search on the validation split that adds questions one at a time to maximize validation PR-AUC, as detailed in Appendix~\ref{app:subset}. The selected set of eight questions differs by model but recurs around a shared core: the local-slur question m1 is selected for all three LLMs, and derogation and context questions appear repeatedly. On the held-out test set, the eight-question model achieves $71.2$--$71.9$ ROC-AUC, within $0.9$ to $1.5$ points of the full twenty-question model at $72.5$--$73.4$, while reducing the number of LLM calls by $60\%$. Figure~\ref{fig:subset-curve} shows performance as a function of the number of selected questions. The selected sets draw on all three question groups, which confirms that boundary and language/context questions are not redundant with content questions.

% \paragraph{Takeaway.}
% \textbf{Across three LLMs, targeted questions predict disagreement far better than asking directly, boundary and context questions add consistent signal beyond harmful content, and eight questions retain nearly all of it.}

% Requires: \usepackage{booktabs,multirow,graphicx}
% Sources: global_module_importance_shap.csv, global_question_importance_shap.csv,
% dataset_model_top5_questions.csv (native RF importance). Provisional.

\subsection{What Signals Disagreement}
\label{sec:res-explanation}

\begin{figure*}[t]
\centering
\includegraphics[width=\linewidth]{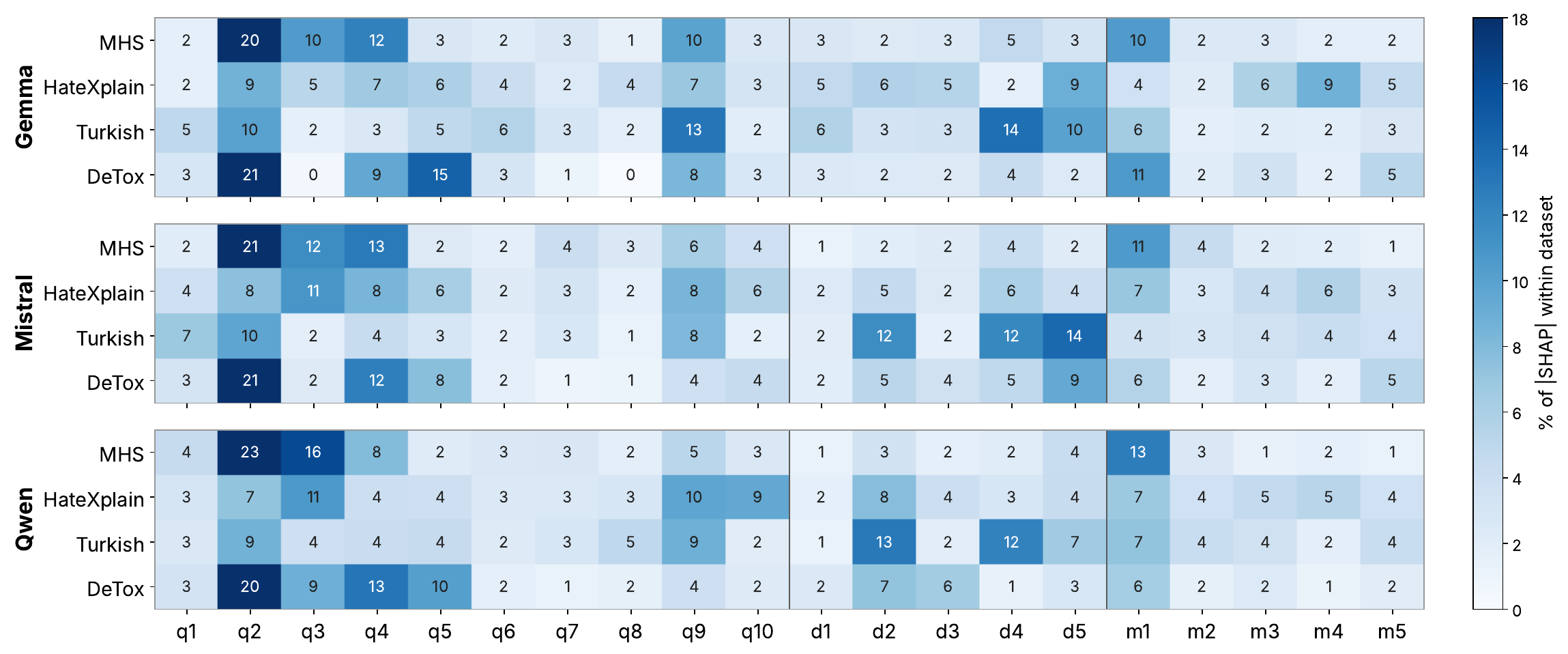}
\caption{Per-question TreeSHAP importance, as each question’s share of the total attribution within a dataset, for the three LLMs. Questions are grouped into content (q), boundary (d), and language/context (m).}
\label{fig:shap-heatmap}
\end{figure*}

\paragraph{A shared core of questions.} We use TreeSHAP~\citep{lundberg2017unifiedapproachinterpretingmodel} to attribute each prediction to the questions, summing a question’s answer and confidence contributions. Two questions recur across datasets and LLMs: the derogation question q2 is among the five most important in all twelve dataset--LLM combinations, and the context-or-intent question q9 in ten. The dehumanization question q4 and the local-slur question m1 follow, each in the top five of eight combinations. Disagreement is signaled not by the presence of harmful content alone, but by whether the text is derogatory, together with cues that its interpretation is uncertain. We illustrate these predictions on individual posts in Appendix~\ref{app:worked-examples}, showing, for three HateXplain items, the questions that most strongly raise or lower the predicted disagreement.

\paragraph{Language-specific cues.} Which questions matter beyond this core depends strongly on the dataset, as Figure~\ref{fig:shap-heatmap} shows. On MHS, content questions dominate, led by derogation (q2, 21.0\%), slurs (q3, 12.7\%), the local slur (m1, 11.2\%), and dehumanization (q4, 11.0\%). On DeTox, derogation (q2, 20.6\%), dehumanization (q4, 11.7\%), and scapegoating (q5, 10.8\%) rank highest. Turkish is the clear exception: the slang-and-coded-language question d4 (12.6\%), borderline severity (d5, 10.2\%), context (q9, 10.0\%), and the abuse-versus-harm boundary (d2, 9.3\%) are its most important questions, each above every content question except derogation (q2, 9.4\%). HateXplain is the flattest, with no question above 8.9\% and content, boundary, and language cues all contributing, consistent with its lower overall predictability. The prominence of d4 and d5 on Turkish supports the view that disagreement in non-English data depends more on language-specific and interpretive cues than in the English datasets.

% \paragraph{Boundary and context questions are efficient.} Content questions are ten of the twenty, yet on Turkish, the five boundary questions together carry more of the attribution than all ten content questions, and on the other datasets, the boundary and language/context groups take a share comparable to their count despite being half as many. On a per-question basis, they are therefore at least as influential as content questions, which matches the predictive gains from adding them in \S\ref{sec:res-prediction}.

\paragraph{Boundary and context questions are efficient.} Content questions are ten of the twenty, yet in Turkish, the five boundary questions carry more of the attribution than all ten content questions, and on the other datasets, the boundary and language/context groups together take a share comparable to their count. On a per-question basis, they are therefore at least as influential as content questions, which matches the predictive gains from adding them in \S\ref{sec:res-prediction}. The prominence of d4 and m1 in Turkish supports the view that disagreement in non-English data depends more on language-specific and interpretive cues than in the English datasets.

\subsection{Transfer Across Datasets}
\label{sec:res-transfer}

\begin{table}[t]
\centering
\small
\setlength{\tabcolsep}{3.2pt}
\begin{tabular}{@{}p{0.2cm}p{0.9cm} p{0.5cm}p{0.5cm}p{0.7cm} p{0.5cm}p{0.5cm}p{0.7cm} p{0.5cm}p{0.5cm}@{}}
\toprule
& & \multicolumn{3}{c}{\textbf{In-domain}} & \multicolumn{3}{c}{\textbf{Transfer}} & \multicolumn{2}{c}{\textbf{Rel.}} \\
\cmidrule(lr){3-5} \cmidrule(lr){6-8} \cmidrule(lr){9-10}
& \textbf{LLM} & ROC & PR & MCC & ROC & PR & MCC & ROC & PR \\
\midrule
\multirow{3}{*}{\rotatebox[origin=c]{90}{\textbf{MHS}}}
& Gemma    & 74.7 & 61.3 & 37.0 & 68.6 & 53.5 & 23.9 & 91.8 & 87.3 \\
& Mistral  & 76.1 & 63.1 & 39.3 & 68.1 & 54.0 & 23.4 & 89.5 & 85.6 \\
& Qwen     & 76.3 & 63.3 & 38.6 & 70.2 & 55.8 & 26.6 & 92.0 & 88.1 \\
\midrule
\multirow{3}{*}{\rotatebox[origin=c]{90}{\textbf{Xplain}}}
& Gemma    & 64.6 & 60.9 & 21.0 & 60.2 & 56.6 & 14.7 & 93.2 & 93.0 \\
& Mistral  & 65.5 & 61.7 & 22.4 & 61.5 & 58.0 & 16.1 & 93.8 & 94.0 \\
& Qwen     & 66.0 & 62.5 & 23.1 & 62.2 & 58.1 & 17.4 & 94.2 & 92.9 \\
\midrule
\multirow{3}{*}{\rotatebox[origin=c]{90}{\textbf{Turkish}}}
& Gemma    & 73.4 & 64.7 & 36.2 & 69.5 & 60.6 & 25.4 & 94.7 & 93.6 \\
& Mistral  & 72.2 & 63.5 & 35.2 & 68.7 & 59.7 & 20.7 & 95.2 & 94.0 \\
& Qwen     & 73.2 & 65.0 & 35.9 & 68.3 & 59.8 & 22.3 & 93.3 & 92.1 \\
\midrule
\multirow{3}{*}{\rotatebox[origin=c]{90}{\textbf{DeTox}}}
& Gemma    & 81.6 & 48.7 & 41.2 & 75.7 & 39.0 & 33.5 & 92.7 & 80.0 \\
& Mistral  & 82.4 & 50.4 & 40.5 & 75.0 & 38.6 & 31.8 & 91.1 & 76.7 \\
& Qwen     & 84.1 & 51.9 & 44.7 & 78.3 & 41.9 & 38.2 & 93.0 & 80.7 \\
\bottomrule
\end{tabular}
\caption{Leave-one-dataset-out transfer. In-domain models are trained and evaluated on the held-out dataset with cross-validation; transfer models are trained on the other three datasets without any explicit language or dataset features. Rel.\ is the ratio of transfer to in-domain performance (\%). All values in \%.}
\label{tab:transfer}
\end{table}

\paragraph{Ranking transfers across datasets.}
Models trained on three datasets and applied to a fourth retain 89.5--95.2\% of in-domain ROC-AUC, as shown in Table~\ref{tab:transfer}, even though the held-out dataset differs in language, source platform, annotation guidelines, label set, and disagreement rate, and no explicit language or dataset feature is available to the model. Transfer is strongest for Turkish, at 93.3--95.2\%, and HateXplain, at 93.2--94.2\%, and lowest for MHS, where it falls to 89.5\% for Mistral. The pattern holds for all three LLMs. Transfer models also outperform direct prompting on the same unseen data by a wide margin, for example, 78.3 against 67.5 ROC-AUC on DeTox and 70.2 against 59.9 on MHS for Qwen, so a predictor transferred from other languages is preferable to asking the LLM directly about the target data.

\paragraph{Ranking transfers better than operating points.}
The cost of transfer is higher for the threshold-dependent metrics. PR-AUC retains 92.1--94.0\% on HateXplain and Turkish, the two datasets with disagreement rates close to one half, but only 76.7--88.1\% on DeTox and MHS, where disagreement is rarer. MCC falls further, for instance, from 35.9 to 22.3 on Turkish and from 44.7 to 38.2 on DeTox for Qwen. Transferred models, therefore, rank contested items reliably, while their scores are shifted with respect to the target distribution, which matters when a fixed threshold is required.

\section{Conclusion}
\label{sec:conclusion}

We studied whether annotator disagreement in harmful content can be predicted from interpretable properties of the text. An LLM answers a fixed set of binary questions about content, boundary cues, and language-specific cues, and a random forest predicts disagreement from the answers. Across four datasets in English, German, and Turkish and three open-weight LLMs, this outperforms asking the same LLMs directly whether annotators would disagree, as well as a supervised text baseline, and a compact set of eight questions retains most of the predictive signal. Which questions matter differ sharply across datasets: content questions dominate on MHS, while boundary questions carry most of the signal on Turkish. Predictors also transfer to unseen datasets and languages, retaining 89.5--95.2\% of in-domain ROC-AUC without any explicit language or dataset feature. Disagreement in harmful content is thus partly predictable from explicit, interpretable properties of the text, and these properties are shared enough across datasets to support transfer.

\section*{Limitations}

We acknowledge that this study has some limitations. 
% First, we do not have human judgments for the questions themselves, since none of the four datasets provides annotations at this level of granularity, and collecting them for three languages was beyond the scope of this work; the consistency of our findings across three independently developed LLMs suggests that the answers capture stable properties of the text, and targeted validation on a sample remains a natural next step. 
First, our study covers three languages, four datasets, and three open-weight LLMs of comparable size, which is broader than prior work on disagreement prediction, and extending it to further languages, subjective tasks, and model scales is straightforward given the dataset-independent design of the questions.

Second, we predict whether annotators disagree rather than how strongly, which is the only target that is comparable across datasets that differ in the number of annotators per item and in whether an intermediate category is available; finer-grained targets, such as soft labels, are available for some datasets and offer a promising extension. 

Third, feature attribution identifies the properties the predictor relies on rather than the reasons annotators disagreed; we therefore report group-level attribution alongside individual questions and read individual ranks with corresponding care. Fourth, we deliberately model disagreement from the text alone, without annotator identities or demographics, which makes the approach applicable to datasets that do not release annotator information; combining text-side and annotator-side signals is complementary and left to future work.

Finally, our design is structural: because predictions are made from a fixed set of named questions, the feature space is human-readable by design, and a prediction can be traced to the questions it relies on. We quantify this reliance with treeSHAP, but we do not claim that these attributions are mechanistically faithful, nor do we use attention-based or other model-internal explanation methods; identifying the questions a predictor uses is not the same as identifying why annotators disagreed.

\section*{Ethical Considerations}

Our work predicts where annotators are likely to disagree on harmful content, and we see two main considerations. First, disagreement scores are intended to route contested items to human review and to guide additional annotation, not to automate moderation decisions or to discard contested items. Down-weighting items that annotators find divisive could suppress minority perspectives, which runs counter to the perspectivist motivation of this work; the scores should support human judgment, not replace it. Second, our features are produced by LLMs, which carry known cultural and demographic biases in harmful-content judgments, so predicted disagreement reflects the models' sensitivities as much as the annotators'. We therefore report results across three model families and four datasets rather than relying on a single model. 

All datasets were used under their original licenses and terms, for their intended research purpose, and we collected no new annotations and processed no personally identifying information beyond what the datasets already contain. The harmful examples in these datasets are offensive by nature, and we handled them only for research on annotation disagreement.

\section*{Acknowledgements}
The work was supported by the European Research Council (ERC) through the European Union's Horizon Europe research and innovation program (grant agreement No. 101113091) and the German Research Foundation (DFG; grant FR 2829/7-1).

\bibliography{custom}
\bibliographystyle{acl_natbib}

\appendix

% \section{Example Appendix}
% \label{sec:appendix}

% Requires in preamble: \usepackage{booktabs,multirow,amsmath,amssymb}
% Placeholders marked with [TODO: ...] must be filled before submission.

\appendix

\section{Dataset Details}
\label{app:datasets}

We use four harmful-content datasets that provide multiple annotations per item, which allows disagreement to be measured. Table~\ref{tab:dataset-details} reports their statistics. An item is counted as a disagreement when its annotators are not unanimous.

MHS (Measuring Hate Speech; \citealp{kennedy2026measuringhatespeechspectrum,sachdeva-etal-2022-measuring}) contains English comments from YouTube, Reddit, and Twitter, annotated by US-based crowdworkers. Each annotation gives one of three responses: no hate speech, unclear, or hate speech. We use these responses directly, keeping the intermediate “unclear” class distinct.~\footnote{Note that “unclear” denotes annotator uncertainty and is not equivalent to the “offensive” class in HateXplain and Turkish.} HateXplain \citep{mathew2022hatexplainbenchmarkdatasetexplainable} contains English posts from Twitter and Gab, each labeled by three annotators as hate, offensive, or normal, following the hate/offensive distinction of \citet{davidson2017automatedhatespeechdetection}. 

Turkish \citep{toraman-etal-2022-large} contains Turkish tweets spanning five domains (religion, gender, race, politics, and sports), each labeled by five annotators as hate, offensive, or normal. The available annotations provide per-class label counts but not annotator identifiers, so annotator-level statistics are not reported for this dataset. DeTox \citep{demus-etal-2022-comprehensive} contains German comments from social media. It provides a binary harmful/non-harmful label, so its disagreement label reflects agreement over a single binary judgment rather than a three-way choice.

\begin{table*}[t]
\centering
\small
\setlength{\tabcolsep}{4pt}
\begin{tabular}{@{}lrrrr@{}}
\toprule
& \textbf{MHS} & \textbf{HateXplain} & \textbf{Turkish} & \textbf{DeTox} \\
\midrule
Language & English & English & Turkish & German \\
License & CC-BY 4.0 & MIT & CC BY-NC-SA 4.0 & GPL-3.0 \\
Label classes & 3 & 3 & 3 & 2 \\
Comments & 29{,}453 & 19{,}192 & 60{,}308 & 9{,}689 \\
Total annotations & 135{,}450 & 57{,}687 & 301{,}540 & 28{,}908 \\
Unique annotators & 7{,}912 & 253 & --- & 6 \\
\midrule
Annotators / comment (min) & 1 & 3 & 5 & 1 \\
Annotators / comment (mean) & 4.60 & 3.01 & 5.00 & 2.98 \\
Annotators / comment (max) & 815 & 3 & 5 & 7 \\
Comments / annotator (min) & 6 & 1 & --- & 41 \\
Comments / annotator (mean) & 17.1 & 228.0 & --- & 3{,}132.4 \\
Comments / annotator (max) & 26 & 5{,}730 & --- & 4{,}972 \\
\midrule
Disagreement rate & 0.404 & 0.487 & 0.466 & 0.215 \\
Krippendorff's $\alpha$ & 0.471 & 0.506 & 0.694 & 0.341 \\
\bottomrule
\end{tabular}
\caption{Dataset statistics. Disagreement rate is the fraction of comments with non-unanimous annotations. Krippendorff's $\alpha$ is the nominal alpha over the label distributions. Annotator identifiers are not available for the Turkish dataset, so annotator-level statistics are omitted.}
\label{tab:dataset-details}
\end{table*}

\section{Models Details}
\label{app:models}
We use three open-weight, instruction-tuned LLMs of comparable size from different families as described in Table~\ref{tab:models}: Gemma-3-27B\footnote{\url{https://huggingface.co/google/gemma-3-27b-it}}~\citep{gemmateam2025gemma3technicalreport}, Mistral-Small-3.2-24B\footnote{\url{https://huggingface.co/mistralai/Mistral-Small-3.2-24B-Instruct-2506}}~\citep{mistral_small_32_2025}, and Qwen-3.6-27B\footnote{\url{https://huggingface.co/Qwen/Qwen3.6-27B}}~\citep{qwen3.6-27b} obtained from HuggingFace and served with vLLM~\citep{kwon2023efficientmemorymanagementlarge}. For Qwen 3.6, the thinking mode is disabled. All three are decoder-only and released under permissive open-weight licenses. We run each model for inference only, with greedy decoding, and restrict generation to the single answer token so that the \emph{yes}/\emph{no} token log probabilities can be read directly for the confidence features.

The disagreement classifier is a random forest with balanced class weights, applied to the question answers and their confidences. It uses 120 trees, a minimum of 8 samples per leaf, a maximum of 0.75 of the training samples per tree, and a fixed seed. A separate classifier is trained for each LLM, so features from different LLMs are never mixed. In-domain results use three outer folds, grouped by normalized text so that no text appears in more than one fold, with an inner validation split used to tune the decision threshold.

\begin{table}[t]
\centering
\small
\setlength{\tabcolsep}{4pt}
\begin{tabular}{@{}lccc@{}}
\toprule
& \textbf{Gemma 3} & \textbf{Mistral S. 3.2} & \textbf{Qwen 3.6} \\
\midrule
Params & 27B & 24B & 27B \\
Developer & Google & Mistral AI & Alibaba \\
License & Gemma & Apache 2.0 & Apache 2.0 \\
Context & 128K & 128K & 262K \\
\bottomrule
\end{tabular}
% TODO: verify Qwen 3.6 27B parameter count, license, and context length against the model card.
\caption{The three LLMs used to answer the questions.}
\label{tab:models}
\end{table}

In-domain results use 3-fold cross-validation, stratified by disagreement label and grouped by item so that no item appears in both training and test folds; for the pooled and subset-selection experiments, the group key is the original item identifier shared across the three LLMs. Baselines and question models are evaluated on identical folds.
% \paragraph{Computational Resources. }
The experiments were carried out on machines with NVIDIA GeForce GTX 1080 Ti GPUs. During evaluation, models were used in inference mode without any parameter updates. 

\section{Questions and Prompts}
\label{app:prompts}

Table~\ref{tab:questions} lists all twenty questions. 

\begin{table*}[t]
\centering
\small
\begin{tabular}{llp{11.5cm}}
\toprule
\textbf{Group} & \textbf{ID} & \textbf{Question} \\
\midrule
\multirow{10}{*}{Content}
 & C1  & Does the text reference a protected group (e.g., based on race, religion, gender, sexual orientation, or origin)? \\
 & C2  & Does the text describe the group negatively or derogatorily? \\
 & C3  & Does the text use slurs or stereotypes? \\
 & C4  & Does the text dehumanize or vilify the group? \\
 & C5  & Does the text blame or scapegoat the group? \\
 & C6  & Does the text call for excluding or discriminating against the group? \\
 & C7  & Does the text threaten the group or wish harm upon it? \\
 & C8  & Does the text directly incite violence or action against the group? \\
 & C9  & Considering context and intent, is the text meant to harm or demean? \\
 & C10 & Is the text likely to have a harmful effect on the target? \\
\midrule
\multirow{5}{*}{Boundary}
 & B1  & Is the target of the text ambiguous or unclear? \\
 & B2  & Is the text on the boundary between general abuse and harm toward a group? \\
 & B3  & Is context missing that would be needed to judge the text? \\
 & B4  & Does the text use slang, sarcasm, irony, or coded language? \\
 & B5  & Is the severity of the text borderline? \\
\midrule
\multirow{5}{*}{Language/context}
 & L1  & Does the text contain a local slur or coded insult specific to a language or community? \\
 & L2  & Does the text refer to a protected group using local or community-specific wording? \\
 & L3  & Does the text invoke a local conflict or intergroup tension? \\
 & L4  & Does the text exclude or disparage the group indirectly through local references? \\
 & L5  & Does the interpretation depend on a local pragmatic cue? \\
\bottomrule
\end{tabular}
\caption{Full question set. Question IDs map to the result tables as C1--C10 = q1--q10, B1--B4 = d1--d4, B5 = d5, and L1--L5 = m1--m5.}
\label{tab:questions}
\end{table*}

\section{Label-Based Control Baselines}
\label{app:label-controls}

To test whether disagreement is explained by the harmful-content label itself, we compare our question-based predictor against two controls that are given the human gold label as input: a random forest trained on the gold majority label, and one trained on the gold label together with the LLM's harmful-content output. Both are \emph{privileged}: the gold majority label is not available for a new, unlabeled text, so these are upper-reference baselines rather than deployable systems. They are trained on the combined data across all datasets, so we report them at that scope in the Table~\ref{tab:label-controls}.

\begin{table}[h]
\centering
\small
\setlength{\tabcolsep}{5pt}
\begin{tabular}{@{}ll ccc@{}}
\toprule
\textbf{LLM} & \textbf{Features} & ROC & PR & F1$_m$ \\
\midrule
\multirow{3}{*}{Gemma 3}
& Gold majority label        & 64.0 & 54.0 & 63.1 \\
& Gold + LLM output          & 72.2 & 61.8 & 66.2 \\
& Question-based (ours)      & \textbf{72.8} & 61.6 & \textbf{66.3} \\
\midrule
\multirow{3}{*}{Mistral}
& Gold majority label        & 64.0 & 54.0 & 63.1 \\
& Gold + LLM output          & 69.1 & 59.1 & 66.2 \\
& Question-based (ours)      & \textbf{72.9} & \textbf{61.6} & \textbf{66.4} \\
\midrule
\multirow{3}{*}{Qwen 3.6}
& Gold majority label        & 64.0 & 54.0 & 63.1 \\
& Gold + LLM output          & 70.2 & 59.9 & 66.6 \\
& Question-based (ours)      & \textbf{73.6} & \textbf{62.6} & \textbf{66.7} \\
\bottomrule
\end{tabular}
\caption{Combined-trained comparison against label-based controls (\%). The gold majority label is a privileged input not available at inference time. ROC and PR are ROC-AUC and PR-AUC; F1$_m$ is macro-F1. Best per block in \textbf{bold}.}
\label{tab:label-controls}
\end{table}

The question-based predictor outperforms the gold-label-only control by 8.8--9.6 ROC-AUC points for all three LLMs, indicating that disagreement is not merely a function of the majority label. It also outperforms the stronger gold-plus-output control on the combined data, despite that baseline's access to the gold label. The one exception is HateXplain, where the label-based control is higher; the near-even disagreement rate makes the aggregated label an especially strong signal.

\section{Fold Variability}
\label{app:std}
Table~\ref{tab:std} reports the fold standard deviations behind the main results, for the full 20-question model and the compact 8-question model. Variability is small on the larger datasets, with ROC-AUC standard deviations below 0.5 points on MHS and Turkish for all three LLMs, and somewhat higher on HateXplain and DeTox, which have fewer test instances per fold. In all cases, the standard deviations are well below the gaps between our models and the baselines in the main table, so the ranking of methods is stable across folds. We omit the standard deviations from the main table for readability.

\begin{table*}[h]\centering\small\setlength{\tabcolsep}{4pt}
\begin{tabular}{@{}ll cc@{}}\toprule
\textbf{LLM} & \textbf{Dataset} & \textbf{Full (20)} & \textbf{Compact (8)} \\
& & ROC\,/\,PR & ROC\,/\,PR \\\midrule
Gemma 3 27B & MHS & 74.7$\pm$0.7\,/\,61.3$\pm$0.9 & 73.8$\pm$0.8\,/\,60.1$\pm$1.5 \\
 & HateXplain & 64.6$\pm$1.2\,/\,60.9$\pm$0.9 & 63.8$\pm$1.1\,/\,60.2$\pm$1.5 \\
 & Turkish & 73.4$\pm$0.2\,/\,64.7$\pm$0.2 & 71.8$\pm$0.3\,/\,62.8$\pm$0.4 \\
 & DeTox & 81.6$\pm$0.1\,/\,48.7$\pm$1.1 & 80.2$\pm$0.4\,/\,46.6$\pm$0.4 \\
\midrule
Mistral Small 3.2 24B & MHS & 76.1$\pm$0.5\,/\,63.1$\pm$0.7 & 75.8$\pm$0.4\,/\,63.2$\pm$0.7 \\
 & HateXplain & 65.5$\pm$1.2\,/\,61.7$\pm$1.1 & 64.9$\pm$1.0\,/\,61.0$\pm$0.6 \\
 & Turkish & 72.2$\pm$0.2\,/\,63.5$\pm$0.2 & 71.2$\pm$0.2\,/\,62.3$\pm$0.2 \\
 & DeTox & 82.4$\pm$0.1\,/\,50.4$\pm$1.4 & 81.2$\pm$0.6\,/\,47.6$\pm$1.7 \\
\midrule
Qwen 3.6 27B & MHS & 76.3$\pm$0.5\,/\,63.3$\pm$0.4 & 76.0$\pm$0.5\,/\,63.4$\pm$0.6 \\
 & HateXplain & 66.0$\pm$1.2\,/\,62.5$\pm$1.1 & 64.7$\pm$1.0\,/\,61.6$\pm$0.9 \\
 & Turkish & 73.2$\pm$0.1\,/\,65.0$\pm$0.1 & 71.7$\pm$0.4\,/\,63.2$\pm$0.4 \\
 & DeTox & 84.1$\pm$0.3\,/\,51.9$\pm$1.0 & 82.5$\pm$0.0\,/\,49.5$\pm$1.3 \\
\bottomrule\end{tabular}
\caption{Mean $\pm$ fold standard deviation (three outer folds) for the full 20-question and compact 8-question confidence models (ROC-AUC\,/\,PR-AUC, \%).}
\label{tab:std}
\end{table*}

\section{Compact Question Subset Selection}
\label{app:subset}

Because each question requires one LLM call per item, a smaller question set lowers inference cost. For each LLM, we select a compact subset with a validation-based beam search, using the same confidence-only features as the full model so that selection and the reported predictor are consistent.

The search considers all twenty questions. The selection objective is validation PR-AUC, with ties broken by validation ROC-AUC, then by fewer questions, then by question ID.

The beam search keeps a beam of the $b=8$ best partial subsets at each size:
(1): Train a random forest on every single question and score it on the validation split.
(2): Retain the $b$ best single questions.
(3): Expand each retained subset by adding one question not already included.
(4): Score each expanded subset on validation and retain the $b$ best.
(5): Repeat until size eight.
(6): Select the subset of size at most eight with the highest validation PR-AUC, and report it on the held-out test split.

Beam search is used because exhaustively evaluating all subsets of size up to eight is computationally prohibitive. The selected sets differ by model: {q2, q7, q8, q9, d5, m1, m2, m3} for Gemma, {q1, q2, q3, d2, m1, m2, m3, m4} for Mistral, and {q1, q3, q4, q5, q9, d3, m1, m2} for Qwen, with the local-slur question m1 selected for all three.

% Requires: \usepackage{booktabs}
% Content warning: the quoted dataset texts contain hateful and offensive language.

\section{Worked Examples}
\label{app:worked-examples}

We walk through three HateXplain instances in English, using the Qwen 3.6 question answers and the full twenty-question random forest trained on HateXplain in Table~\ref{tab:worked-index}. For each one, we list the five questions with the largest attributions by magnitude, where the attribution is the signed TreeSHAP contribution to the disagreement probability relative to the model's reference value of $0.50$, and a positive value pushes the probability up. These contributions describe the trained model, not the annotators' reasons, and they are not causal effects: a question can carry a positive attribution while its answer is \emph{no}, so an attribution does not confirm that the property is present.~\footnote{\textcolor{red}{Warning: these examples contain hateful language.}}

\begin{table}[h]
\centering\small\setlength{\tabcolsep}{4pt}
\begin{tabular}{@{}llcccc@{}}
\toprule
& \textbf{Votes} & \textbf{Target} & \textbf{RF $P$} & \textbf{RF} & \textbf{Direct} \\
\midrule
E1 & 3 hatespeech               & 0 & 0.163 & 0 & 0 \\
E2 & 1 normal / 2 offensive     & 1 & 0.695 & 1 & 0 \\
E3 & 2 normal / 1 offensive     & 1 & 0.297 & 0 & 1 \\
\bottomrule
\end{tabular}
\caption{The three worked instances: annotator votes, disagreement target, random forest disagreement probability and prediction at threshold $0.49$, and the direct-prompting prediction. }
\label{tab:worked-index}
\end{table}

\paragraph{E1: agreed-upon hate.}
All three annotators label E1 hatespeech, so the target is agreement; the random forest predicts $0.163$ and direct prompting also predicts agreement.

\textit{you are a kike loving nigger who still uses your mom insults you should just do the world a favor and attach yourself to the end of a rope}

\begin{table}[h]\centering\small\setlength{\tabcolsep}{5pt}
\begin{tabular}{@{}lp{3.3cm}cc r@{}}
\toprule
\textbf{Q} & \textbf{Dimension} & \textbf{Ans.} & \textbf{Conf.} & \textbf{SHAP} \\
\midrule
d2 & Abuse vs.\ hate boundary        & no  & 1.00 & $-0.073$ \\
d3 & Missing context                 & no  & 1.00 & $-0.048$ \\
q3 & Hateful slurs or stereotypes    & yes & 1.00 & $-0.041$ \\
d5 & Borderline severity             & no  & 1.00 & $-0.037$ \\
q2 & Negative/derogatory description & yes & 1.00 & $-0.029$ \\
\bottomrule
\end{tabular}
\caption{Feature attributions for example E1.}\label{e1}
\end{table}
As shown in Table~\ref{e1}, for this sample, the content questions fire, but the boundary, missing-context, and severity questions are all answered \emph{no}, and their negative attributions keep the probability low, so unambiguous hate that annotators agree on is correctly not predicted as disagreement.

\paragraph{E2: contested boundary case.}
E2 is labeled normal by one annotator and offensive by two, so the target is disagreement; the random forest predicts $0.695$ while direct prompting predicts agreement and is wrong.

\textit{when i first got on here and said i hate trump i was considered a troll but people who yell about hating niggers are not trolls yeah ok}

\begin{table}[h]\centering\small\setlength{\tabcolsep}{5pt}
\begin{tabular}{@{}llcc r@{}}
\toprule
\textbf{Q} & \textbf{Dimension} & \textbf{Ans.} & \textbf{Conf.} & \textbf{SHAP} \\
\midrule
m5 & Local pragmatic cue          & yes & 0.99 & $+0.027$ \\
q10 & Likely effect on target     & yes & 1.00 & $+0.023$ \\
d2 & Abuse vs.\ hate boundary     & yes & 0.92 & $+0.023$ \\
m4 & Indirect local exclusion     & no  & 0.98 & $+0.021$ \\
q3 & Hateful slurs or stereotypes & yes & 1.00 & $-0.018$ \\
\bottomrule
\end{tabular}
\caption{Feature attributions for example E2.}\label{e2}
\end{table}
For this sample, according to Table~\ref{e2}, the abuse-versus-hate boundary question is answered \emph{yes} and is among the strongest positive drivers, which is the kind of near-boundary signal that separates contested items from clear ones and that direct prompting misses.

\paragraph{E3: reclaimed term, a failure case.}
E3 is labeled normal by two annotators and offensive by one, so the target is disagreement, but the random forest predicts $0.297$, below threshold, while direct prompting is correct.

\textit{that nigga kawhi is nice bruh}

\begin{table}[h]\centering\small\setlength{\tabcolsep}{5pt}
\begin{tabular}{@{}llcc r@{}}
\toprule
\textbf{Q} & \textbf{Dimension} & \textbf{Ans.} & \textbf{Conf.} & \textbf{SHAP} \\
\midrule
q3 & Hateful slurs or stereotypes  & yes & 1.00 & $-0.053$ \\
m4 & Indirect local exclusion      & no  & 1.00 & $-0.041$ \\
q10 & Likely effect on target      & no  & 0.80 & $-0.030$ \\
q9 & Context and intent            & no  & 0.98 & $-0.028$ \\
m3 & Historical/political conflict & no  & 1.00 & $-0.015$ \\
\bottomrule
\end{tabular}
\caption{Feature attributions for example E3.}\label{e3}
\end{table}
\noindent
The slur question is answered \emph{yes} yet contributes negatively (Table~\ref{e3}), because the surrounding context questions are answered \emph{no}, so the model treats the reclaimed term as friendly and under-predicts the disagreement that the split votes show.

\section{AI Usage Statement}
The authors used an AI assistant to support coding and writing. For code, it helped draft and debug analysis and evaluation scripts, which the authors reviewed, tested, and verified
against the reported results. For writing, it assisted with phrasing and copy-editing of text drafted by the authors. All research ideas, experimental design, analysis, and conclusions are the authors' own, and the authors take full responsibility for the content of the paper.

\end{document}